\documentclass[10pt,twocolumn,letterpaper]{article}

\usepackage{wacv}              
\usepackage[accsupp]{axessibility}

\definecolor{wacvblue}{rgb}{0.21,0.49,0.74}
\usepackage[pagebackref,breaklinks,colorlinks,allcolors=wacvblue]{hyperref}

\usepackage{microtype}      
\usepackage{url}            
\usepackage{comment}        

\usepackage{graphicx}
\usepackage[table,dvipsnames]{xcolor} 

\usepackage{amsmath}
\usepackage{amssymb}
\usepackage{amsfonts}       
\usepackage{bm}             
\usepackage{nicefrac}       
\usepackage{pifont}         

\usepackage{booktabs}       
\usepackage{multirow}
\usepackage{makecell}

\usepackage{wrapfig}
\usepackage{caption}
\usepackage{subcaption}

\usepackage{algorithm}
\usepackage{algorithmic}

\definecolor{Gray}{gray}{0.95}
\newcolumntype{g}{>{\columncolor{Gray}}c}

\def\paper{VIOLA}
\definecolor{NavyBlue}{rgb}{0.0, 0.0, 0.5}

\def\wacvPaperID{830} 
\def\confName{WACV}
\def\confYear{2027}

\title{\paper: Towards Video In-Context Learning \\ with Minimal Annotations} 

\author{Ryo Fujii$^{1,2}$ \hspace{0.6cm} Hideo Saito$^{1,2}$  \hspace{0.6cm} Ryo Hachiuma$^{3}$ \\
$^{1}$Keio University \hspace{0.6cm} $^{2}$Keio AI Research Center \hspace{0.6cm} $^{3}$NVIDIA \\
{\tt\small \{ryo.fujii0112, hs\}@keio.jp, rhachiuma@nvidia.com}\\
}
\begin{document}
\maketitle

\begin{abstract}
Generalizing Multimodal Large Language Models (MLLMs) to novel video domains is essential for real-world deployment but remains challenging due to the scarcity of labeled data. While In-Context Learning (ICL) offers a training-free adaptation path, standard methods rely on large annotated pools, which are often impractical in specialized environments like industrial or surgical settings since they require the experts' annotations. To bridge this gap, we introduce \textbf{\paper} (\textbf{V}ideo \textbf{I}n-c\textbf{O}ntext \textbf{L}earning with minimal \textbf{A}nnotation), a label-efficient framework that synergizes minimal expert supervision with abundant unlabeled data. First, to maximize the efficiency of a strict annotation budget, we propose density-uncertainty-weighted sampling. Unlike standard diversity or uncertainty strategies that risk selecting visual outliers, our method leverages density estimation to identify samples that are simultaneously diverse, representative, and informative. Second, to utilize the remaining unlabeled data without noise propagation, we construct a hybrid pool and introduce confidence-aware retrieval and confidence-aware prompting. These mechanisms explicitly model label reliability, retrieving demonstrations based on a composite score of similarity and confidence while enabling the MLLM to adaptively distinguish between verified ground truths and noisy pseudo-labels. Extensive experiments across nine diverse benchmarks using three MLLMs demonstrate that our framework significantly outperforms various baselines in low-resource settings, achieving robust adaptation with minimal annotation costs. Project page: \url{https://fujiry0.github.io/VIOLA-project-page/}
\end{abstract}

\section{Introduction}
\label{sec:intro}

\begin{figure}[tb]
\centering
\includegraphics[width=\linewidth]{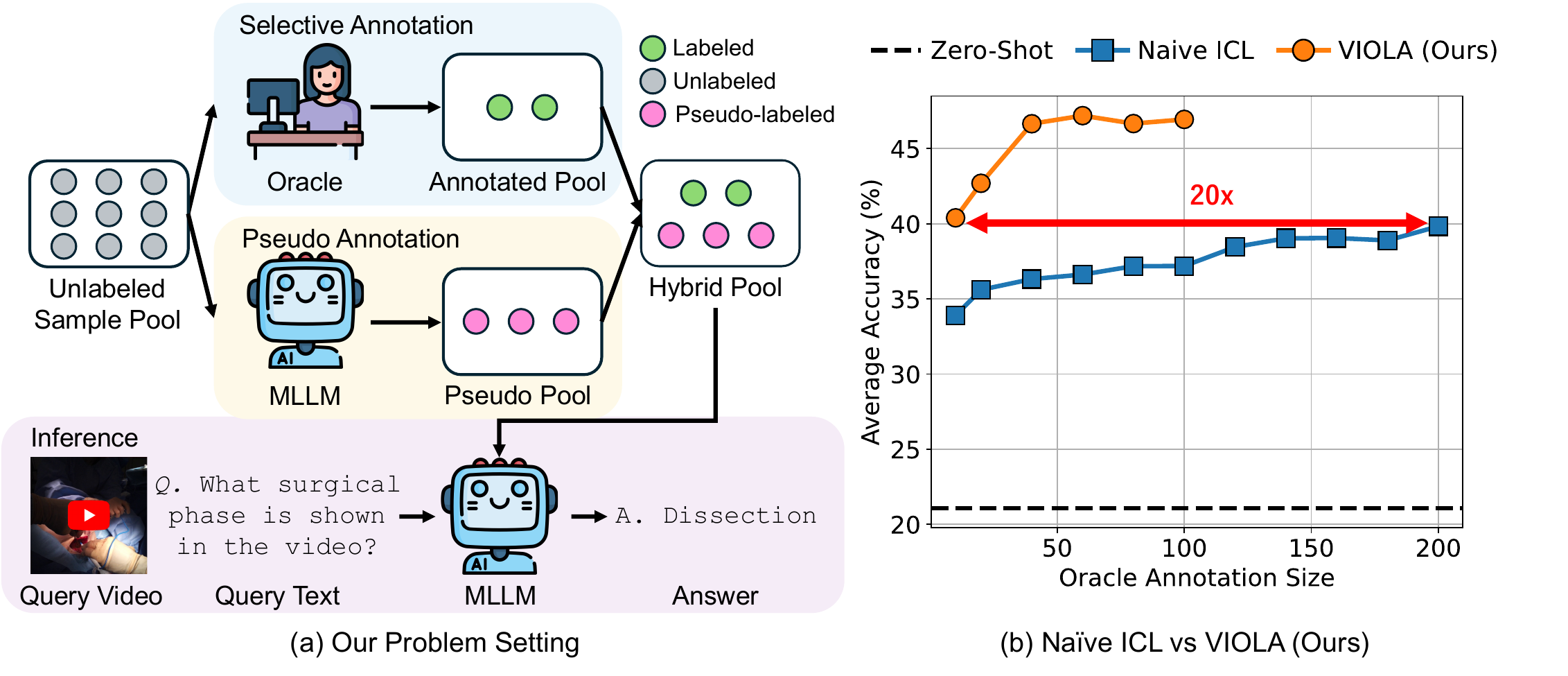}
\caption{\textbf{Problem setup and performance comparison.} (a) Unlike methods requiring large labeled datasets, our approach, VIOLA, strategically selects a minimal subset of informative samples for expert annotation while leveraging abundant unlabeled videos via pseudo-labeling. This constructs a hybrid pool from which the model retrieves relevant demonstrations for inference. (b) Performance comparison with Naive ICL (which randomly selects samples for annotation), averaged across seven classification datasets. VIOLA achieves robust adaptation with significantly reduced annotation costs.}
\label{fig:teaser}
 \vspace{-1.5em}
\end{figure}

Recent advances in Multimodal Large Language Models (MLLMs)~\cite{damonlpsg2025videollama3,Qwen2-VL,lin2024videollava} have notably enhanced video understanding capabilities. However, despite the growing power of these foundational models, novel domains, scenarios, and tasks continuously emerge in real-world applications~\cite{Gabeff2025MammAlps,Bar2024EgoPet,Suriya2017Xsports,Martin2019drive,Bashmal2023CapERA,sun2024bora,sultani2018Crime,wang2025medgenunlockingmedicalvideo}. Consequently, relying solely on pre-trained knowledge is insufficient; models must possess the capability to adapt to these unseen data distributions. While fine-tuning on target videos offers a straightforward path to enhance performance, it is often impractical since it requires substantial training data to avoid overfitting and incurs high computational costs (\eg, GPU memory), making it difficult to scale across continuously emerging scenarios. In contrast, In-Context Learning (ICL)~\cite{Brown2020ICL} has emerged as an efficient alternative. Originally pioneered in Large Language Models (LLMs), ICL allows models to adapt to new domains by simply providing example inputs (\ie, demonstrations) alongside a test sample during inference, bypassing the need for weight parameter updates via backpropagation. This paradigm has demonstrated strong generalization on unseen tasks in language-only~\cite{Brown2020ICL, xie2022an, rubin2022learning} and image-language domains~\cite{alayrac2022Flamingo, li2023otter}, and is now being extended to the video-language domain~\cite{Kim2025VideoICL} as a powerful strategy for low-resource adaptation.

A critical bottleneck for ICL in the video domain is the difficulty of obtaining high-quality labeled demonstrations. This issue is exacerbated in specialized or emerging domains such as industrial~\cite{Ragusa2024ENIGMA} and surgical~\cite{fujii2024egosurgery,he2024pitvqa} footage, as interpreting such content demands high-level expert knowledge, substantially increasing annotation costs compared to general scenarios. Consequently, the standard assumption in existing ICL methods~\cite{Kim2025VideoICL}, that a large pool of annotated examples is readily available for retrieval, is often untenable in these novel environments. The labor-intensive nature of video annotation, combined with the necessity for expert involvement, creates a substantial barrier to deploying ICL in real-world video applications where rapid adaptation to new data is essential.

To address this realistic and practical challenge, ICL research leverages unlabeled data via two different strategies: selective annotation~\cite{su2023selective,mavromatis2024covericl,zhang2024ideal,zhang2022active,chen2024fastgas,Yu2023Patron}, which curates specific examples for labeling from an unlabeled sample pool, and pseudo-annotation~\cite{gu2025IterPSD,chen2025maple,mamooler2025picle,berger2025context}, which uses model predictions to transform unlabeled data into demonstrations. While these strategies have proven effective individually, the synergistic potential of combining selective annotation with pseudo-annotation remains largely underexplored. In this work, we explore this avenue by integrating these sources into a hybrid pool as shown in~\Cref{fig:teaser} (a), developing a label-efficient framework that bridges minimal human supervision with extensive unlabeled videos, achieving robust adaptation with significantly reduced annotation costs as demonstrated in~\Cref{fig:teaser} (b).

However, transposing these strategies to the video domain presents three unique challenges: 

\noindent \textbf{Inefficacy of selection:} Current selective annotation methods typically prioritize diversity~\cite{su2023selective,zhang2023automatic} or representativeness~\cite{chen2024fastgas,zhang2024ideal}. Highlighting diversity in data selection is crucial for comprehensive coverage, but may sacrifice representativeness by overlooking exemplary data. This trade-off is particularly critical in the video domain. Unlike text, video features are prone to outliers due to high redundancy and task-irrelevant noise (\eg, background variations). Selecting such outliers merely for the sake of diversity results in demonstrations that contribute minimally to downstream performance, as they fail to provide valid guidance during inference. Consequently, diversity and representativeness need to be balanced carefully to avoid selecting outlier samples that are diverse but not representative.

\noindent \textbf{Indiscrimination in hybrid retrieval:} To address data scarcity, our framework operates on a hybrid pool comprising a small set of expert-annotated ground truths and a large volume of pseudo-labeled videos. However, standard retrieval mechanisms rely solely on similarity~\cite{Kim2025VideoICL,alayrac2022Flamingo,qin2024context,liu2022makes,zhao2024multi-modal}, implicitly assuming uniform label quality. This assumption is unrealistic because reliability varies significantly not only between ground truths and pseudo-labels but also within the pseudo-labeled subset itself. Consequently, a retrieval system driven purely by similarity is prone to selecting visually close instances that may carry erroneous labels, regardless of whether they originate from human experts or model predictions, or the associated confidence of those predictions. To mitigate this, robust in-context learning requires a mechanism that strictly balances visual similarity with label reliability, ensuring that the limited context slots are occupied by demonstrations that are both semantically relevant and trustworthy.

\noindent \textbf{Sensitivity to pseudo-label quality:} Unlike the text domain~\cite{gu2025IterPSD,chen2025maple,mamooler2025picle,berger2025context}, where models can leverage many-shot learning~\cite{agarwal2024manyshot} to statistically mitigate label noise, video inputs incur prohibitive token costs. This constraint confines video ICL to a strict few-shot regime. In this setting, the model becomes hypersensitive to the quality of each demonstration, where a single erroneous pseudo-label can induce significant negative transfer and severely degrade reasoning performance. Thus, simply retrieving examples without explicitly modeling their uncertainty leads to brittle performance.

To overcome these limitations, we propose \textit{\paper}, a label-efficient framework that assembles a robust hybrid demonstration pool by integrating minimal human annotations with extensive unlabeled videos.
First, to guarantee diversity while mitigating the risk of selecting semantic outliers, we introduce \textit{density-uncertainty-weighted selection}. This strategy leverages Gaussian mixture models (GMM) to partition the semantic space into distinct clusters. Within each cluster, we identify the most valuable sample by maximizing a selection score that weights model uncertainty with probability density. This ensures that the expert-annotated examples are simultaneously diverse, representative of their semantic group, and highly informative. These labeled samples are then combined with refined pseudo-labeled data to form the hybrid pool.
Second, to resolve indiscrimination in using this mixed data, we propose \textit{confidence-aware retrieval}. This mechanism retrieves demonstrations based on a composite score of visual similarity and label confidence, ensuring that limited context slots are occupied by reliable instances rather than merely visually similar ones.
Finally, addressing hypersensitivity to noise, we design \textit{confidence-aware prompting}. We explicitly encode the source and confidence scores into the prompt, enabling the MLLM to discern the reliability of each demonstration and adaptively weight its reliance on ground-truth versus pseudo-labeled examples for robust reasoning.

Our main contributions are summarized as follows:
\begin{enumerate}
    \item We propose \paper, a label-efficient framework for video in-context learning that bridges the gap between minimal expert supervision and abundant unlabeled data by constructing a confidence-aware hybrid demonstration pool.
    
    \item We introduce density-uncertainty-weighted selection, which integrates GMM-based clustering with density-weighted uncertainty. This strategy ensures diverse data coverage while filtering out semantic outliers, guaranteeing that the limited annotation budget is invested in samples that are simultaneously representative and highly informative.
    
    \item We develop a confidence-aware retrieval mechanism that optimizes the trade-off between visual similarity and label reliability within the hybrid pool, preventing noisy pseudo-labels from degrading the quality of the retrieved context.
    
    \item We design a confidence-aware prompting strategy that explicitly embeds confidence scores into the input, enabling the MLLM to adaptively calibrate its reasoning based on the reliability of the provided demonstrations (ground truths vs. pseudo-labels).
\end{enumerate}

\section{Related Work}
\noindent \textbf{In-Context Learning.}
In-Context Learning (ICL)~\cite{Brown2020ICL, xie2022an,rubin2022learning} empowers models to adapt to novel tasks by conditioning generation on a few input-output demonstrations, thereby eliminating the need for parameter updates. This paradigm has been successfully extended to various Computer Vision (CV) domains, spanning images~\cite{bar2022visual, wang2023images, wang2023sggpt, zhao2024multi-modal, bai2024sequential, qin2023unicontrol, zhang2024instruct, wang2025explore, lee2025cropper,zhao2024mmicl,jiang2024manyshot,Baldassini2024whatmaskemultimodal,alayrac2022Flamingo}, video~\cite{alayrac2022Flamingo,li2023otter, Kim2025VideoICL}, point clouds~\cite{fang2024explore, shao2024micas}, skeleton sequences~\cite{wang2024skeleton-in-context}, pedestrian trajectories~\cite{fujii2025towards}, and 3D hand reconstruction~\cite{Xie2026EgoHandICL}. Notably, VideoICL~\cite{Kim2025VideoICL} recently advanced out-of-distribution video understanding through similarity-based example selection and confidence-based iterative inference.
However, a fundamental limitation of standard video ICL frameworks~\cite{Kim2025VideoICL} is their reliance on large-scale annotated pools for retrieval. Such a prerequisite is impractical in specialized domains~\cite{Ragusa2024ENIGMA,fujii2024egosurgery}, where the prohibitive cost of expert annotation renders the construction of extensive support sets infeasible. In contrast to prior works predicated on abundant supervision, we target the realistic label-efficient setting, optimizing performance under a strict ground-truth expert annotation budget.

\noindent \textbf{Selective Annotation.}
Traditionally, Active Learning (AL)~\cite{NeurIPS1994AL} aims to incrementally annotate samples that maximize model performance with minimal labeling costs. Decades of research have yielded a wide array of strategies to select informative samples~\cite{sener2018coreset,Ash2020BADGE,ash2021BAIT,ParvanehCVPRAlphaMix,mahmood2022lowbudget,yehuda2022active,Xie2023activeft,Li2024Noisy}. In the context of ICL, this paradigm is adapted as selective annotation, optimizing the labeling budget by identifying the most valuable examples from an unlabeled pool. Prior works typically prioritize either diversity~\cite{su2023selective,zhang2023automatic,chen2024fastgas} or uncertainty~\cite{mavromatis2024covericl,zhang2024ideal}. However, directly applying these strategies to the video domain is suboptimal due to the prevalence of noise.
Methods maximizing diversity often inadvertently capture outliers, which are samples that are statistically distinct but semantically unrepresentative due to task-irrelevant noise (\eg, background clutter).
Conversely, uncertainty-based selection assumes that high entropy implies informational value. In multimodal settings, however, this often conflates epistemic uncertainty (useful difficulty) with aleatoric noise (\eg, visual ambiguity), leading to the selection of low-quality demonstrations.
To address this, we propose density-uncertainty-weighted sampling. By synergizing semantic density with model uncertainty, our method explicitly filters noisy outliers while targeting samples that are both representative and informative.

\noindent \textbf{Pseudo-Labeling.}
Pseudo-labeling~\cite{lee2013Pseudo-Label} serves as a cornerstone of semi-supervised learning~\cite{Yang2023SSLSurvey}. While traditional Semi-Supervised Learning (SSL) relies on self-training paradigms~\cite{Xie2020UDA,wang2023freematch,SohnNeurIPSFixMatch,xu2021Dash,chen2023softmatch,Li2021CoMatch,berthelot2022adamatch,Wang2022DebiasPL,NeurIPS2022DebiasPL} leveraging high-confidence predictions, recent studies have adapted this principle to ICL to mitigate data scarcity. Specifically, these works employ pseudo-annotation to scale demonstrations via model predictions~\cite{gu2025IterPSD,chen2025maple,mamooler2025picle,berger2025context}. Such methods show promise in text domains, where models can utilize many-shot learning~\cite{agarwal2024manyshot} to statistically average out label noise. Conversely, video ICL operates under a strict few-shot regime due to prohibitive token costs. In this setting, models become hypersensitive to noise, as a single erroneous demonstration can induce significant negative transfer. Current retrieval mechanisms~\cite{li2023finding,shum2023automatic}, relying primarily on similarity~\cite{Kim2025VideoICL,alayrac2022Flamingo,qin2024context,liu2022makes,zhao2024multi-modal}, fail to distinguish between ground truths and potentially noisy pseudo-labels. In contrast, our framework introduces confidence-aware mechanisms spanning both retrieval and prompting to explicitly model label reliability, enabling robust inference even with noisy hybrid pools.

Distinct from prior studies that treat selective annotation and pseudo-labeling in isolation, our approach is grounded in the Semi-Supervised Active Learning (SSAL) paradigm~\cite{Yin2025toward,gao2020consistency,Guo2024IDEAL,Huang2021TOD,Li2020ASCENT,xie2025multi}. SSAL merges the strengths of AL and SSL, aiming to significantly improve model performance under conditions of limited annotation. By adapting this paradigm to the video ICL context, we propose a unified label-efficient framework that strategically synergizes minimal expert supervision with abundant unlabeled data via pseudo-annotation. This holistic and unified approach ensures robust adaptation and maximizes performance in realistic low-resource target domains.

\begin{figure*}[tb]
\centering
\includegraphics[width=0.9\linewidth]{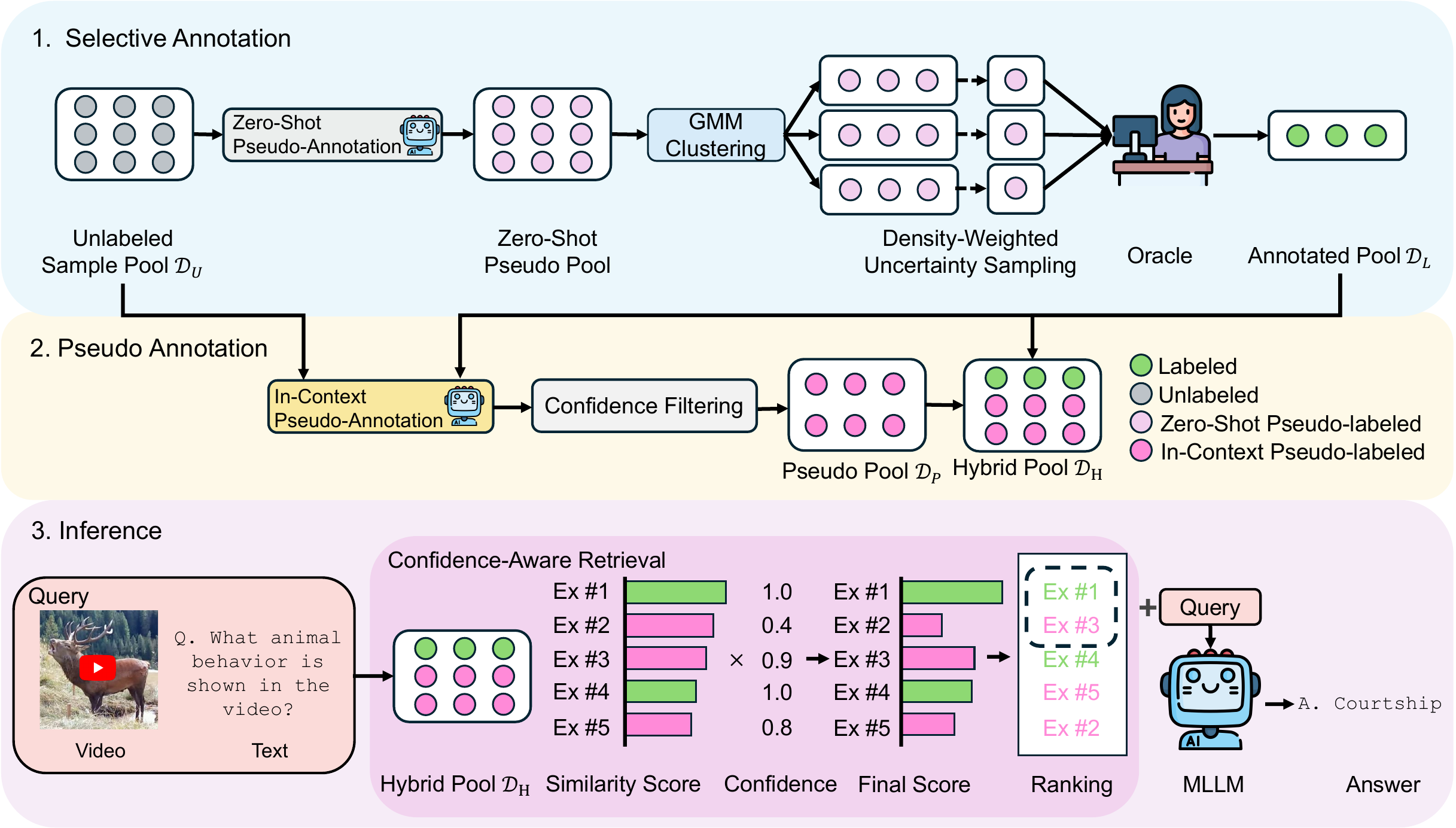}
\caption{
  \textbf{Overview of our proposed framework.}
  The pipeline consists of three stages. First, we acquire expert labels for a small, informative subset ($\mathcal{D}_L$) using density-uncertainty-weighted sampling (Selective annotation stage). Then, we generate high-confidence pseudo-labels via in-context pseudo-annotation to construct a hybrid pool $\mathcal{D}_H$ (Pseudo-annotation stage). Lastly, we predict final answers via confidence-aware retrieval and prompting during inference.
  }
\label{fig:overview}
 \vspace{-1.5em}
\end{figure*}

\section{Methodology}
\label{sec:method}
\subsection{Preliminaries}
\label{sec:preliminaries}
In the standard In-Context Learning (ICL) paradigm, a pre-trained Multimodal Large Language Model (MLLM), denoted as $\mathcal{M}$, and a demonstration pool $\mathcal{D} = \{(x_i, y_i)\}_{i=1}^{N}$ are typically employed. Here, $x_i$ represents the multimodal input (video and text query) and $y_i$ is the corresponding ground-truth answer.

Given a test query $x_{test}$, the inference process involves retrieving a specific context set $\mathcal{C}$ comprising $m$ demonstrations that are semantically relevant to the query. Typically, this retrieval is based on visual similarity within the latent space~\cite{Kim2025VideoICL,alayrac2022Flamingo,zhao2024multi-modal}. The index set $\mathcal{I}$, corresponding to the top-$m$ samples with the highest similarity scores, is identified as:
\begin{equation}
    \mathcal{I} = \operatorname*{arg\,top\textit{-}m}_{i \in \{1, \dots, N\}} \text{sim}(x_{test}, x_i),
\end{equation}
where $\text{sim}(\cdot, \cdot)$ denotes the cosine similarity between visual embeddings. Based on these indices, the in-context prompts are constructed as $\mathcal{C} = \{(x_i, y_i) \mid i \in \mathcal{I}\}$. Finally, the model predicts the target answer $\hat{y}$ by conditioning on both the retrieved context and the test query:
\begin{equation}
    \hat{y} = \mathcal{M}(x_{test}, \mathcal{C}).
\end{equation}
A key assumption in this standard setting is that the pool $\mathcal{D}$ is fully annotated and sufficiently large to cover the distribution of $x_{test}$.

\subsection{Overview}
\label{sec:overview}
In this work, we address a realistic and challenging scenario where expert annotations are scarce. To overcome this limitation, our framework operates on a hybrid demonstration pool $\mathcal{D}_H$, constructed by integrating a small set of expert-annotated ground-truth samples with a large volume of pseudo-labeled videos.
We start with a pool of unlabeled instances $\mathcal{D}_U = \{u_i\}_{i=1}^{N}$, where each instance $u_i$ consists of a video and a corresponding task query. We assume a strict annotation budget $B$ ($B \ll N$), representing the maximum number of samples an expert can label.

We assemble this hybrid pool $\mathcal{D}_H$ by strategically allocating the budget and leveraging the remaining unlabeled data, ultimately enabling robust inference. As illustrated in ~\cref{fig:overview}, this process consists of three key stages.

\noindent \textbf{Selective Annotation Stage.} 
We first aim to select an optimal subset of indices $\mathcal{S} \subset \{1, \dots, N\}$ such that $|\mathcal{S}| = B$. An oracle (human expert) provides labels for these selected samples, forming the labeled set:
\begin{equation}
    \mathcal{D}_L = \{(u_i, y_i) \mid i \in \mathcal{S}\}.
\end{equation}
To maximize the efficiency of this budget, we propose density-uncertainty-weighted selection. As detailed in ~\cref{sec:selective_annotation}, this strategy synergizes semantic density with model uncertainty to identify samples that are diverse, representative, and informative.

\noindent \textbf{Pseudo Annotation Stage.} 
To utilize the vast amount of remaining unlabeled data $\mathcal{D}_{rem} = \mathcal{D}_U \setminus \mathcal{D}_L$, we employ the model $\mathcal{M}$ to generate pseudo-labels while utilizing $\mathcal{D}_L$ as contexts. Crucially, since video ICL is sensitive to noise, we also estimate a confidence score $c_j$ for each sample. This results in a pseudo-labeled set:
\begin{equation}
    \mathcal{D}_P = \{(u_j, \hat{y}_j, c_j) \mid u_j \in \mathcal{D}_{rem}\},
\end{equation}
where $(\hat{y}_j, c_j) \leftarrow \mathcal{M}(u_j, \mathcal{C}_{\mathcal{D}_L})$. We describe the specifics of this in-context pseudo-annotation process in \cref{sec:pseudo_annotation}.

\noindent \textbf{Inference Stage.} 
Finally, the hybrid demonstration pool is defined as $\mathcal{D}_H = \mathcal{D}_L \cup \mathcal{D}_P$. For a test query $x_{test}$, we retrieve a context set $\mathcal{C}$ from $\mathcal{D}_H$ to enable robust inference:
\begin{equation}
    \hat{y} = \mathcal{M}(x_{test}, \mathcal{C}).
\end{equation}
To mitigate the impact of noisy pseudo-labels, we introduce confidence-aware retrieval and confidence-aware prompting, which are elaborated in ~\cref{sec:inference}.

\subsection{Selective Annotation via Density-Uncertainty-weighted Selection}
\label{sec:selective_annotation}
In the first phase, our objective is to identify the optimal subset $\mathcal{S}$ for expert annotation. To maximize the utility of the limited budget $B$, we propose density-uncertainty-weighted sampling. This strategy addresses the limitations of diversity-only or uncertainty-only methods by synergizing semantic density with model uncertainty to filter out outliers while targeting informative samples.

\noindent \textbf{Semantic Density Estimation.}
We first capture the underlying semantic structure of the unlabeled pool $\mathcal{D}_U$. We extract video embeddings $z_i$ and fit a Gaussian Mixture Model (GMM) with $K$ components, where $K$ is set equal to the budget $B$. For each sample $u_i$, we compute the posterior probability $\gamma_{ik}$ of belonging to component $k$:
\begin{equation}
    \gamma_{ik} = \frac{\pi_k \mathcal{N}(z_i \mid \mu_k, \Sigma_k)}{\sum_{j=1}^{K} \pi_j \mathcal{N}(z_i \mid \mu_j, \Sigma_j)}.
\end{equation}
A high $\gamma_{ik}$ indicates that the sample is representative of the $k$-th semantic cluster.

\noindent \textbf{Zero-Shot Uncertainty Estimation.}
Concurrently, we estimate prediction uncertainty to gauge the informativeness of each sample. We prompt the frozen MLLM $\mathcal{M}$ in a zero-shot manner. The confidence score $c^{zero}_i$ is defined as the minimum token probability in the generated sequence:
\begin{equation}
    c^{zero}_i = \min_{t=1}^{T} P(w_t | u_i),
\end{equation}
where $w_t$ denotes the $t$-th token in the generated sequence, and $T$ represents the total sequence length.

\noindent \textbf{Density-uncertainty-weighted Sampling.}
To ensure diverse coverage while targeting informative samples, we enforce a stratified policy. For each cluster $k$, we define a selection score that weights uncertainty with density:
\begin{equation}
    S_{k}(u_i) = \gamma_{ik}^{(1-\lambda)} \cdot (1 - c^{zero}_i)^\lambda,
\label{eq:lambda}
\end{equation}
where $\lambda \in [0, 1]$ is a hyperparameter that balances the trade-off between representativeness and informativeness. A higher $\lambda$ prioritizes uncertain samples (hard examples), while a lower $\lambda$ favors high-density samples (prototypical examples) to avoid outliers.
We select exactly one instance from each cluster that maximizes this score to form the labeled set $\mathcal{D}_L$. This guarantees that the demonstrations are diverse (covering all $K$ modes), representative, and informative.

\subsection{In-Context Pseudo-Annotation}
\label{sec:pseudo_annotation}
Once the expert-labeled set $\mathcal{D}_L$ is obtained, we proceed to the pseudo annotation phase to utilize the remaining unlabeled data $\mathcal{D}_{rem} = \mathcal{D}_U \setminus \mathcal{D}_L$. A critical insight here is that while zero-shot signals were sufficient for selection, they are too noisy to serve as demonstrations. Therefore, we employ an in-context pseudo-annotation strategy.

\noindent \textbf{Pseudo Annotation.}
To generate high-quality pseudo-labels, we leverage the expert knowledge encapsulated in $\mathcal{D}_L$. For each unlabeled sample $u_j \in \mathcal{D}_{rem}$, we retrieve a local context set $\mathcal{C}_j \subset \mathcal{D}_L$. Since $\mathcal{D}_L$ consists exclusively of expert-annotated ground truths, we apply the standard retrieval mechanism based on visual similarity. We then perform few-shot inference to obtain a refined label $\hat{y}_j$ and a confidence score $c_j$ derived from the output token probability:
\begin{equation}
    (\hat{y}_j, c_j) \leftarrow \mathcal{M}(u_j, \mathcal{C}_j).
\end{equation}
Unlike the zero-shot estimation used in the selection phase, this step utilizes in-context examples to align the prediction with the specific domain semantics. To strictly control the quality of the generated data, we filter these predictions based on their confidence. Specifically, we retain only samples where $c_j$ exceeds the 95th percentile of the confidence distribution, ensuring that only the most reliable pseudo-labels are added to the pool.

\noindent \textbf{Hybrid Pool Construction.}
We aggregate these processed samples to form the pseudo-labeled pool $\mathcal{D}_P = \{(u_j, \hat{y}_j, c_j)\}$. Finally, to enable the retrieval of both expert-verified and model-generated demonstrations during inference, we unite the two sources into a single hybrid pool:
\begin{equation}
    \mathcal{D}_H = \mathcal{D}_L \cup \mathcal{D}_P.
\end{equation}
This hybrid pool serves as the foundation for the subsequent confidence-aware inference stage.

\subsection{Confidence-Aware Inference}
\label{sec:inference}

In the final phase, we leverage the hybrid pool $\mathcal{D}_H$ to generate answers for test queries. To mitigate the risk of noise propagation from pseudo-labels, we introduce two confidence-aware mechanisms.

\noindent \textbf{Confidence-Aware Retrieval.}
As discussed in ~\cref{sec:preliminaries}, standard retrieval relies solely on visual similarity ($\operatorname*{arg\,top\textit{-}m} \text{sim}$), implicitly treating expert annotations and pseudo-labels as equally reliable. To rectify this, we propose a retrieval score $r$ that penalizes uncertain pseudo-labels. For a candidate $u_i$ in the hybrid pool $\mathcal{D}_H$, the score is computed as:
\begin{equation}
    r_i = \text{sim}(u_i, x_{test})^{(1-\tau)} \cdot (c_i)^\tau,
\end{equation}
where $\tau$ balances semantic relevance with reliability. For ground-truth samples ($u_i \in \mathcal{D}_L$), we set $c_i=1.0$, ensuring they are prioritized if semantically relevant.
Consequently, we retrieve the context indices $\mathcal{I}$ by maximizing this composite score:
\begin{equation}
    \mathcal{I} = \operatorname*{arg\,top\textit{-}m}_{u_i \in \mathcal{D}_H} r_i.
\end{equation}
This formulation ensures that the limited context slots are occupied by demonstrations that optimize the trade-off between visual similarity and label trustworthiness.

\noindent \textbf{Confidence-Aware Prompting.}
To explicitly communicate reliability, we apply a formatting function $\Phi$ that
adapts based on the demonstration source. For a retrieved demonstration
$(u_k, y_k, c_k)$ where $k \in \mathcal{I}$, the refined answer
$\tilde{y}_k = \Phi(y_k, c_k)$ embeds a machine-readable confidence marker into
$y_k$, distinguishing curated ground-truth demonstrations ($u_k \in \mathcal{D}_L$)
from confidence-scored pseudo-labels ($u_k \in \mathcal{D}_P$):
\begin{equation}
    \tilde{y}_k =
    \begin{cases}
        \Phi_{\text{gt}}(y_k) & \text{if } u_k \in \mathcal{D}_L, \\
        \Phi_{\text{pseudo}}(y_k, c_k) & \text{if } u_k \in \mathcal{D}_P.
    \end{cases}
\end{equation}
This differentiation acts as a soft-gating mechanism, allowing the MLLM to attend
strongly to verified patterns while cautiously integrating high-confidence
pseudo-labels. The exact prompt templates are provided in the 
Supplementary.

Finally, we construct the context set $\mathcal{C}$ by pairing the input instances
with their formatted answers for all retrieved indices:
\begin{equation}
    \mathcal{C} = \{(u_k, \tilde{y}_k) \mid k \in \mathcal{I}\}.
\end{equation}
With this refined context, we perform the final inference:
\begin{equation}
    \hat{y} = \mathcal{M}(x_{test}, \mathcal{C}).
\end{equation}

\section{Experiments} 
\subsection{Experimental Settings}
\noindent \textbf{Datasets.} We validate our framework across two video-language tasks using nine diverse datasets spanning medical, industrial, ego-centric, and surveillance domains. For video classification, we employ accuracy as the evaluation metric across seven benchmarks: Drive\&Act~\cite{Martin2019drive} for driver activity recognition; EgoPet~\cite{Bar2024EgoPet} for animal ego-centric interaction recognition; EgoSurgery~\cite{fujii2024egosurgery,Fujii2022Surgicaltool,Fujii2024Egosurgerytool} for medical procedure analysis; ENIGMA~\cite{Ragusa2024ENIGMA} for industrial human-object interaction recognition; UCF-Crime~\cite{sultani2018Crime} for surveillance anomaly detection; Xsports~\cite{Suriya2017Xsports} for extreme sports recognition; and MammAlps~\cite{Gabeff2025MammAlps} for animal behavior recognition. For video captioning, we adopt the ROUGE-L score using Bora~\cite{sun2024bora} for the biomedical domain and CapERA~\cite{Bashmal2023CapERA} for aerial surveillance scenes. Details regarding prompts and class definitions are provided in the Supplementary Material.

\noindent \textbf{Baselines.} (1) \textbf{Zero-Shot} provides only the task instruction to the MLLM, without any demonstrations. (2) \textbf{Random} performs random example selection for annotation. (3) \textbf{Random+Pseudo Annotation} uses the MLLM to generate pseudo-labels for the unlabeled instances to serve as additional annotated data. (4) \textbf{Random+VideoICL}~\cite{Kim2025VideoICL} utilizes a confidence-based iterative inference mechanism to ensure high-quality generation. Note that while the original method assumes a fully labeled pool, we implement it using a randomly selected pool in our experiments. (5) \textbf{VoteK}~\cite{su2023selective} selects representative examples from $B$ buckets stratified by the model's confidence scores.

\noindent \textbf{Implementation Details.}
We evaluate our framework using several state-of-the-art open-source MLLMs, specifically Qwen2-VL-7B~\cite{Qwen2-VL}, VideoLLaMA3-7B~\cite{damonlpsg2025videollama3}, and Qwen3-VL-8B~\cite{Qwen3-VL}. These models were selected for their superior performance on general video understanding benchmarks. For feature extraction, we utilize InternVideo2~\cite{wang2024internvideo2} as the video encoder. Videos are sampled at 1 frame per second (FPS); for sequences exceeding 32 seconds, we uniformly downsample to 32 frames to maintain temporal consistency. In our in-context learning setup, we set the number of retrieved demonstrations to $m=8$, where each demonstration video consists of 16 frames. Furthermore, we apply a quality control step by filtering all generated pseudo-demonstrations using a confidence threshold at the 95th percentile. Regarding prompt arrangement, pseudo-labeled demonstrations are positioned at the beginning of the context, followed by labeled demonstrations. Within each group, examples are sorted by similarity to the test sample, placing the most relevant ground-truth demonstrations next to the target query.

\begin{table*}[tb]
    \caption{\textbf{Main results on diverse video understanding benchmarks.} We evaluate our framework against various baselines across classification and captioning tasks, using a fixed pool of 20 labeled samples for each dataset. $\Delta$ denotes the performance gain of Ours relative to the Zero-shot baseline. Results for VideoICL are reproduced using the official repository\protect\footnotemark.} 
    \centering
    \resizebox{\textwidth}{!}{%
    \begin{tabular}{l|lccccccccc}
        \toprule
        \multicolumn{1}{c}{} & \multirow{2.5}{*}{Methods}   &  \multicolumn{7}{c}{Classification} &   \multicolumn{2}{c}{Captioning} 
        \\ 
        \cmidrule(l{2pt}r{2pt}){3-9} \cmidrule(l{2pt}r{2pt}){10-11}
        \multicolumn{1}{c}{}  &  & Drive$\&$Act & EgoPet & EgoSurgery & ENIGMA & UCF-Crime & Xsports & MammAlps & Bora & CapERA   \\ \midrule
         \multirow{7}{*}{\rotatebox{90}{Qwen2-VL}}
         & Zero-shot & 7.2 & 14.7 & 25.5 & 4.9 & 31.5 & 18.5 & 43.8 & 0.261 & 0.274 \\
         & Random    & 20.9 & 19.0 & 26.8 & \textbf{58.8} & 37.9 & 21.5 & 64.2 & 0.338 & 0.378 \\
         & ~~+Pseudo Annotation  & 15.4 & 32.4 & 21.8 & 58.2 & 39.3 & 8.0 & 46.4 & 0.257 & 0.312 \\ 
         & ~~+VideoICL~\cite{Kim2025VideoICL}  & 11.2 & 16.2 & 28.3 & 57.5 & 34.7 & 23.5 & 63.2 & 0.330 & 0.335 \\
         & VoteK~\cite{su2023selective}  & 18.1 & 21.2 & 13.2 & 54.8 & 36.3 & \textbf{26.5} & 64.4 & 0.359 & 0.382 \\
          & \cellcolor{gray!20} Ours & \cellcolor{gray!20}\textbf{26.4} & \cellcolor{gray!20}\textbf{52.9} & \cellcolor{gray!20}\textbf{30.1} & \cellcolor{gray!20}58.5 &
          \cellcolor{gray!20}\textbf{41.1} & \cellcolor{gray!20}25.9 & \cellcolor{gray!20}\textbf{65.0} &
          \cellcolor{gray!20}\textbf{0.365}&
          \cellcolor{gray!20}\textbf{0.393}\\
          & \cellcolor{gray!20}$\Delta$ & \cellcolor{gray!20}\textcolor{teal}{+19.2} & \cellcolor{gray!20}\textcolor{teal}{+38.2} & \cellcolor{gray!20}\textcolor{teal}{+4.6} & \cellcolor{gray!20}\textcolor{teal}{+53.6} &  
          \cellcolor{gray!20}\textcolor{teal}{+9.6}  & \cellcolor{gray!20}\textcolor{teal}{+7.4} & \cellcolor{gray!20}\textcolor{teal}{+21.2} &
          \cellcolor{gray!20}\textcolor{teal}{+0.104} &
          \cellcolor{gray!20}\textcolor{teal}{+0.119}\\
         \midrule
         \multirow{7}{*}{\rotatebox{90}{VideoLLaMA3}}
         & Zero-shot & 21.7 & 10.7 & 24.0 & 44.3 & 31.7 & 20.2 & 56.4 & 0.220 & 0.228 \\
         & Random   & \underline{36.0} & \underline{16.1} & \underline{38.3} & 47.6 & 34.0 & 21.3 & 67.5 & 0.311 & 0.382 \\
         & ~~+Pseudo Annotation  & 28.3 & 14.1 & 25.6 & \underline{50.5} & 27.4 & 22.8 & 58.4 & 0.224 & 0.297 \\
         & ~~+VideoICL~\cite{Kim2025VideoICL}  & 31.8 & 15.8 & 35.5 & 47.5 & \underline{34.5} & 19.9 & 68.0 & 0.283 & 0.325 \\
         & VoteK~\cite{su2023selective}  & 34.5 & 12.9 & 33.3 & 44.3 & 33.3 & \textbf{31.3} & \underline{68.5} & \textbf{0.325} & \textbf{0.400} \\
         & \cellcolor{gray!20} Ours & \cellcolor{gray!20}\textbf{37.5} & \cellcolor{gray!20}\textbf{31.0} & \cellcolor{gray!20}\textbf{42.5} & \cellcolor{gray!20} \textbf{52.6} & 
          \cellcolor{gray!20} \textbf{36.9} &
          \cellcolor{gray!20} \underline{27.9}  & \cellcolor{gray!20}  \textbf{68.7} &
          \cellcolor{gray!20} \underline{0.313} &
          \cellcolor{gray!20} \underline{0.384}\\ 
         & \cellcolor{gray!20}$\Delta$ & \cellcolor{gray!20}\textcolor{teal}{+15.8} & \cellcolor{gray!20}\textcolor{teal}{+20.3} & \cellcolor{gray!20}\textcolor{teal}{+18.5} & \cellcolor{gray!20}\textcolor{teal}{+8.3} & \cellcolor{gray!20}\textcolor{teal}{+5.2}
          & \cellcolor{gray!20}\textcolor{teal}{+7.7} & \cellcolor{gray!20}\textcolor{teal}{+12.3} & \cellcolor{gray!20}\textcolor{teal}{+0.093}
           & \cellcolor{gray!20} \textcolor{teal}{+0.156} \\
          \midrule
         \multirow{7}{*}{\rotatebox{90}{Qwen3-VL}}
         & Zero-shot & 22.1 & 32.1 & 22.8 & 53.8 & 35.7 & 29.9 & 55.2 & 0.243 & 0.235 \\
         & Random    & \underline{34.8} & 32.8 & 36.7 & 65.6 & 36.9 & 29.9 & 61.4 & \underline{0.276}  & 0.372 \\
         & ~~+Pseudo Annotation  & 29.7 & 30.8 & 25.9 & \underline{66.5} & \underline{37.5} & 29.0 & 60.5 &  0.239 & 0.256 \\
         & ~~+VideoICL~\cite{Kim2025VideoICL}  & 34.3 & 32.4 & \underline{37.1} & 65.9 & \underline{37.5} & 28.2 & 60.9 & 0.253 & 0.297 \\
         & VoteK~\cite{su2023selective}  & 34.5 & \underline{38.4} & 35.9 & 63.5 & 31.0 & \textbf{32.5} & \underline{61.6} & 0.265 & \underline{0.373} \\
          & \cellcolor{gray!20} Ours & \cellcolor{gray!20}\textbf{35.8} & \cellcolor{gray!20}\textbf{48.4} & \cellcolor{gray!20}\textbf{37.7} & \cellcolor{gray!20}\textbf{68.1} & \cellcolor{gray!20}\textbf{39.9} & \cellcolor{gray!20}\underline{31.1} & \cellcolor{gray!20}\textbf{63.7}& \cellcolor{gray!20}\textbf{0.282}& \cellcolor{gray!20}\textbf{0.379} \\
          & \cellcolor{gray!20}$\Delta$ & \cellcolor{gray!20}\textcolor{teal}{+13.7} & \cellcolor{gray!20}\textcolor{teal}{+16.3} & \cellcolor{gray!20}\textcolor{teal}{+14.9} & \cellcolor{gray!20}\textcolor{teal}{+14.3} & \cellcolor{gray!20}\textcolor{teal}{+4.2} & \cellcolor{gray!20}\textcolor{teal}{+1.2} & \cellcolor{gray!20}\textcolor{teal}{+8.5} & \cellcolor{gray!20}\textcolor{teal}{+0.039} & \cellcolor{gray!20}\textcolor{teal}{+0.144} \\
          \bottomrule
    \end{tabular}
    } 
    \label{tab:main}
     \vspace{-1.em}

\end{table*}
\footnotetext{\url{https://github.com/KangsanKim07/VideoICL}}

\subsection{Main Results}
\label{sec:main_results}

~\cref{tab:main} presents a comparison between our method and baseline approaches on nine diverse benchmarks, with an annotation budget of $|\mathcal{D}_L|=20$. From the results, we observe the following phenomenon.

\noindent \textbf{Substantial Improvement over Zero-shot Baselines.} Our method outperforms the zero-shot baseline across all evaluated datasets and architectures. We observe significant gains in specialized domains using Qwen2-VL-7B, such as a $+53.6\%$ increase on the industrial ENIGMA dataset and $+38.2\%$ on the animal-centric EgoPet dataset. These margins highlight that our framework effectively bridges the domain gap where standard zero-shot inference typically struggles, successfully adapting models to unseen distributions without parameter updates.

\noindent \textbf{Superiority over Selective Annotation and Pseudo Annotation Baselines.} In our experimental setting, restricted to a pool of only $20$ labeled samples, the choice of selection strategy becomes critical. We observe that VideoICL, originally designed for full-dataset scenarios, exhibits sensitivity to this limited pool size, often underperforming the simple Random baseline (\eg, $11.2\%$ vs. $20.9\%$ on Drive\&Act with Qwen2-VL-7B). Similarly, Pseudo Annotation frequently yields subpar results due to noisy supervision (\eg, dropping to $8.0\%$ on Xsports). While VoteK offers a more robust baseline, our method still achieves distinct margins over it in classification tasks (\eg, $+31.7\%$ on EgoPet). This advantage extends to captioning tasks in many cases. On the biomedical Bora dataset, our framework achieves an ROUGE-L score of $0.365$ with Qwen2-VL, surpassing both the Random baseline of $0.338$ and the competitive VoteK method of $0.359$. Similarly, on CapERA, we consistently outperform baselines, demonstrating that our confidence-aware retrieval effectively filters noisy pseudo-labels that otherwise degrade the quality of generated descriptions.

\noindent \textbf{Consistent Efficacy across Models.} The effectiveness of our framework is not limited to a specific model. As shown in ~\cref{tab:main}, we observe consistent performance gains when applying our method to VideoLLaMA3-7B and the larger Qwen3-VL-8B. Notably, on VideoLLaMA3-7B, our method secures the best performance in nearly all classification tasks, with gains of up to $+20.3\%$ on EgoPet. Even on the stronger Qwen3-VL-8B model, our approach continues to yield positive improvements. This demonstrates that our framework generally complements capable models by providing high-quality, relevant in-context demonstrations.

\begin{figure*}[tb] 
\centering
\includegraphics[width=0.9\linewidth]{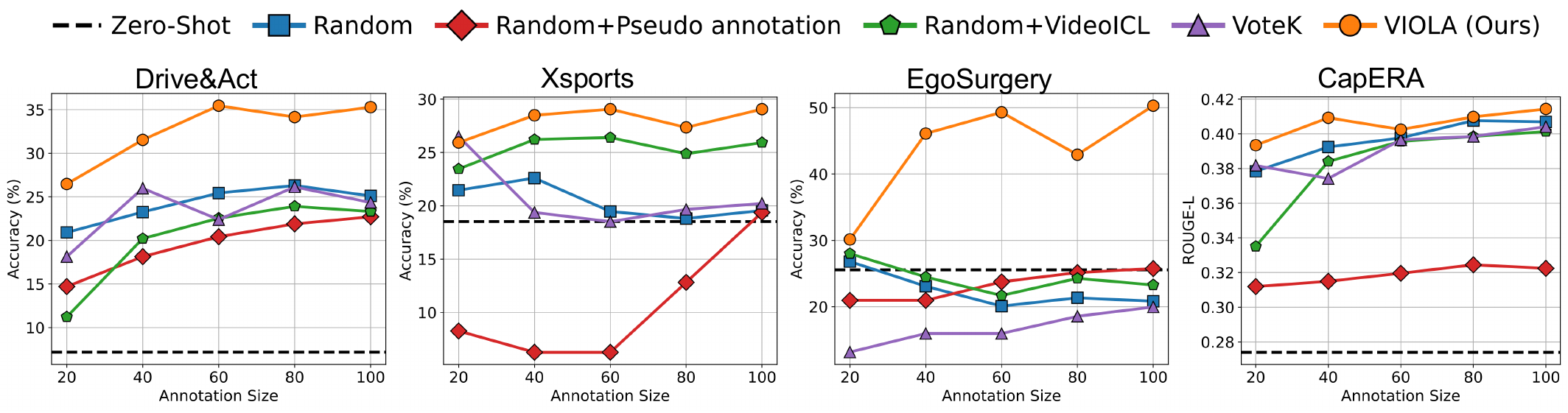}
\caption{\textbf{Performance trends under varying oracle annotation budgets.} We compare our framework against baselines using Qwen2-VL-7B, varying the labeled pool size from 20 to 100 samples.}
\label{fig:qwen2vl-more-labels}
\vspace{-1.em}
\end{figure*}

\subsection{Performance with Varied Budget}
To investigate the scalability of our framework, we evaluate performance variations as the annotation budget $B$ increases from 20 to 100 samples. \cref{fig:qwen2vl-more-labels} illustrates the performance trends using Qwen2-VL. We observe that VideoICL consistently underperforms the Random baseline across most datasets, with the exceptions of EgoSurgery and Xsports, indicating its ineffectiveness within a limited budget setting. In contrast, our method demonstrates remarkable stability, generally outperforming the Random and VoteK baselines regardless of the annotation budget size. Furthermore, our approach yields continuous gains without premature plateauing and maintains a steady advantage in captioning tasks, confirming that our framework scales robustly as more annotations become available.

\subsection{Ablation Studies}
Unless otherwise specified, all ablation studies are performed with an annotation budget of $|\mathcal{D}_L|=20$, aligning with the main results in~\cref{sec:main_results}.

\begin{table}[t]
    \centering
    \caption{Ablation study of density-uncertainty-weighted Sampling.}
    \label{tab:ablation-sampling}
    \resizebox{\linewidth}{!}{%
    \begin{tabular}{lccc >{\columncolor[gray]{0.95}}c ccc}
        \toprule
        $\lambda$ & 0.0 & 0.1  & 0.3  & 0.5  & 0.7  & 0.9  & 1.0 \\
        \midrule 
        Drive\&Act &  24.3 & \textbf{27.4} & 26.7 & 26.4 & 26.3 & 24.5 & 23.0\\
        ENIGMA     &  57.0 & 58.2 & 58.2 & \textbf{58.5} & 56.0 & 56.9 & 56.9\\
        XSports    & 22.2 & 23.9 & 23.9 & \textbf{25.9} & 23.9 & 21.7 & 16.5 \\
        CapEra     &  0.368 & \textbf{0.393} & \textbf{0.393} & \textbf{0.393} & 0.382 & 0.373  & 0.368 \\
        \bottomrule
    \end{tabular}%
    }
    \vspace{-1em}
\end{table}

\begin{table}[t]
    \centering
    \caption{Ablation study of Selective Annotation via Density-Uncertainty-weighted Selection and Confidence-Aware (CA) Inference mechanisms.}
    \label{tab:ablation-inference}
    \resizebox{\linewidth}{!}{%
    \begin{tabular}{c cccccc}
        \toprule
        Sel. Anno. & CA Retrieval & CA Prompt & Drive\&Act & XSports & Bora & CapEra   \\
        \midrule 
                   &              &           & 20.9 & 18.5 & 0.338 & 0.378  \\ 
                   & \checkmark   & \checkmark & 21.9 & 15.1 & 0.335 & 0.372 \\ 
        \checkmark & \checkmark   &           & 17.6 & 15.0 & 0.364 & 0.373 \\
        \checkmark &              & \checkmark & 24.3 & 22.2 & 0.362 & 0.368 \\ 
        \midrule
        \rowcolor[gray]{0.95} \checkmark & \checkmark & \checkmark & \textbf{26.4} & \textbf{25.9} & \textbf{0.365} & \textbf{0.393} \\
        \bottomrule
    \end{tabular}%
    }
\end{table}

\noindent \textbf{Effectiveness of Density-Uncertainty-weighted Selection.}
\Cref{tab:ablation-sampling} validates our balanced selection strategy in \cref{eq:lambda}. Relying solely on uncertainty ($\lambda=1.0$) severely degrades performance (\eg, -9.4\% on XSports) by selecting noisy outliers. Conversely, using only density ($\lambda=0.0$) fails to capture informative signals required for complex reasoning. By synergizing both ($\lambda=0.5$), our method consistently achieves optimal or near-optimal performance across datasets, confirming that demonstrations must be simultaneously representative and informative.

\noindent \textbf{Effectiveness of Confidence-Aware Inference.}
\Cref{tab:ablation-inference} demonstrates the synergy between confidence-aware retrieval and prompting during inference. While individual components provide limited or unstable gains, combining them yields robust improvements across all datasets. This confirms that effective video ICL requires both filtering the hybrid pool for reliability and explicitly guiding the model to distinguish between ground truths and pseudo-labels.

\begin{table}[t]
    \centering
    \caption{Sensitivity analysis of confidence thresholds for Confidence-Aware (CA) mechanisms.}
    \label{tab:ablation-threshold}
    \resizebox{\linewidth}{!}{%
        \begin{tabular}{cccccccccccccc}
            \toprule
            \multirow{2}{*}{CA Retrieval} & \multirow{2}{*}{CA Prompt} & \multicolumn{6}{c}{Drive\&Act} & \multicolumn{6}{c}{CapEra} \\
            \cmidrule(l{2pt}r{2pt}){3-8} \cmidrule(l{2pt}r{2pt}){9-14}
             & & 0 &  50 & 75 & 90 & 95 & 99 & 0 &  50 & 75 & 90 & 95 & 99 \\
            \midrule
             & & 18.7 & 22.7 & 23.6 & 14.5 & 16.1 & 17.4 & 0.356 & 0.358 & 0.362 & 0.362 & 0.368 & 0.371 \\ 
            \midrule
            \checkmark & & 18.9 & 18.1 & 18.1 & 18.6 & 17.6 & 17.4 & 0.372 & 0.373 & 0.373 & 0.373 & 0.373 & 0.372 \\
             & \checkmark  & 25.1 & 25.9 & 26.1 & 23.2 & 24.3 & 24.5 & 0.370 & 0.372 & 0.374 & 0.363 & 0.368 & 0.371 \\
            \midrule
            \rowcolor[gray]{0.95} \checkmark & \checkmark &  \textbf{26.3} & \textbf{26.3} & \textbf{26.3}  & \textbf{26.3} & \textbf{26.4} & \textbf{25.4} & \textbf{0.393}  & \textbf{0.393}  & \textbf{0.393}  & \textbf{0.390}  & \textbf{0.393}  & \textbf{0.393} \\
            \bottomrule
        \end{tabular}
        \vspace{-1.5em}
    }
\end{table}

\noindent \textbf{Sensitivity Analysis of Confidence Thresholds.}
We evaluate the model's robustness to pseudo-label noise across various confidence thresholds ranging from 0\% to 99\% (with a default of 95\%). As shown in \cref{tab:ablation-threshold}, VIOLA achieves consistently high and stable performance throughout the entire range, especially when both CA components are active. This clearly demonstrates its strong resilience against label noise in the pseudo-annotation pool.

\noindent \textbf{Effectiveness of In-Context Pseudo-Annotation.}
\cref{tab:ablation-pseudo} demonstrates the benefit of conditioning pseudo-label generation on expert examples. Unlike the zero-shot baseline, which struggles with domain nuances, our In-Context strategy leverages retrieved expert data to produce higher-quality labels. This consistently improves downstream performance (\eg, $+1.7\%$ on Xsports and $+0.028$ ROUGE-L on Bora), confirming that expert guidance is essential for constructing a reliable hybrid pool.

\begin{table}[t]
    \centering
    \caption{Comparison of pseudo-annotation strategies. We assess the benefit of using in-context examples (In-Context) over raw model predictions (Zero-shot) for generating pseudo-labels.}
    \label{tab:ablation-pseudo}
    \resizebox{0.75\linewidth}{!}{%
        \begin{tabular}{lcc}
            \toprule
            Pseudo-Annotation Strategy & Xsports  & Bora \\
            \midrule
            Zero-shot & 24.2 & 0.337 \\
            \rowcolor{gray!15} \textbf{In-Context (Ours)} & \textbf{25.9}  & \textbf{0.365} \\
            \bottomrule
        \end{tabular}%
        \vspace{-2em}
    }
\end{table}

\section{Conclusion}
\noindent \textbf{Conclusion.}
We present \paper, a unified framework for label-efficient video In-Context Learning that bridges minimal expert supervision with abundant unlabeled data. To balance diversity and representativeness in video selection, we propose density-uncertainty-weighted sampling, which filters semantic outliers while identifying informative samples for annotation. To mitigate sensitivity to noise in hybrid pools, we introduce in-context pseudo-annotation followed by confidence-aware retrieval and prompting. These mechanisms explicitly distinguish verified ground truths from potentially noisy pseudo-labels. Experiments on diverse benchmarks, including specialized medical and industrial domains, demonstrate that our approach outperforms state-of-the-art baselines, validating the need for synergizing strategic data selection with confidence-aware inference.

\noindent \textbf{Limitations and Future Work.}
Our reliance on pre-trained visual embeddings may compromise retrieval accuracy under significant distribution shifts in highly specialized domains. Future work will explore integrating domain-adapted encoders to enhance robustness in such scenarios.

\section{Acknowledgements}
This work was partially supported by JSPS KAKENHI Grant Number 25K03143.

{
    \small
    \bibliographystyle{ieeenat_fullname}
    \bibliography{main}
}

\end{document}